\documentclass[letterpaper]{article} 
\usepackage[]{aaai25}  
\usepackage{times}  
\usepackage{helvet}  
\usepackage{courier}  
\usepackage[hyphens]{url}  
\usepackage{graphicx} 
\usepackage{natbib}  
\usepackage{caption} 
\usepackage{newfloat}
\usepackage{listings}
\DeclareCaptionStyle{ruled}{labelfont=normalfont,labelsep=colon,strut=off} 
\usepackage{subcaption}
\usepackage{enumitem}
\usepackage{amsmath}
\usepackage{amsfonts}
\usepackage{booktabs}
\usepackage{xspace}
\usepackage{multirow}
\usepackage{bm}
\usepackage{xcolor}
\usepackage{soul}
\usepackage{colortbl}
\usepackage{pifont}
\usepackage{makecell}

\usepackage{algorithm}
\usepackage{algorithmic}

\title{TimeCAP: Learning to Contextualize, Augment, and Predict Time Series Events with Large Language Model Agents}
\author{
    Geon Lee\textsuperscript{\rm 1}, 
    Wenchao Yu\textsuperscript{\rm 2},
    Kijung Shin\textsuperscript{\rm 1},
    Wei Cheng\textsuperscript{\rm 2},
    Haifeng Chen\textsuperscript{\rm 2}
}
\affiliations{
    \textsuperscript{\rm 1}KAIST\\
    \textsuperscript{\rm 2}NEC Labs\\
    \{geonlee0325, kijungs\}@kaist.ac.kr, \{wyu, weicheng, haifeng\}@nec-labs.com
}

\begin{document}

\maketitle

\newcommand\red[1]{\textcolor{red}{#1}}
\newcommand\blue[1]{\textcolor{blue}{#1}}

\newcommand{\naive}{\textsc{TimeCP}\xspace}
\newcommand{\method}{\textsc{TimeCAP}\xspace}

\newtheorem{trm}{\textbf{Theorem}}

\definecolor{pink}{RGB}{231, 190, 186}
\newcommand{\hlpink}[1]{{\sethlcolor{pink}\hl{#1}}}

\definecolor{purple}{RGB}{210, 194, 225}
\newcommand{\hlpurple}[1]{{\sethlcolor{purple}\hl{#1}}}

\definecolor{yellow}{RGB}{253, 244, 215}
\newcommand{\hlyellow}[1]{{\sethlcolor{yellow}\hl{#1}}}

\definecolor{orange}{RGB}{234, 208, 191}
\newcommand{\hlorange}[1]{{\sethlcolor{orange}\hl{#1}}}

\definecolor{green}{RGB}{207, 218, 198}
\newcommand{\hlgreen}[1]{{\sethlcolor{green}\hl{#1}}}

\definecolor{lightblue}{RGB}{196, 204, 223}
\newcommand{\hllightblue}[1]{{\sethlcolor{lightblue}\hl{#1}}}

\begin{abstract}
Time series data is essential in various applications, including climate modeling, healthcare monitoring, and financial analytics. 
Understanding the contextual information associated with real-world time series data is often essential for accurate and reliable event predictions.
In this paper, we introduce TimeCAP, a time-series processing framework that creatively employs Large Language Models (LLMs) as contextualizers of time series data, extending their typical usage as predictors.
TimeCAP incorporates two independent LLM agents: one generates a textual summary capturing the context of the time series, while the other uses this enriched summary to make more informed predictions.
In addition, TimeCAP employs a multi-modal encoder that synergizes with the LLM agents, enhancing predictive performance through mutual augmentation of inputs with in-context examples. 
Experimental results on real-world datasets demonstrate that TimeCAP outperforms state-of-the-art methods for time series event prediction, including those utilizing LLMs as predictors, achieving an average improvement of 28.75\% in F1 score. 
\end{abstract}

\section{Introduction}
\label{sec:intro}
Time series data is fundamental to numerous applications, including climate modeling~\cite{schneider1974climate}, energy management~\cite{liu2023sadi}, healthcare monitoring~\cite{liu2023large}, and finance analytics~\cite{sawhney2020deep}.
Consequently, a range of advanced techniques has been developed to capture complex dynamic patterns intrinsic to time series data~\cite{wu2021autoformer,nie2022time,zhang2022crossformer}.
However, real-world time series data often involves essential contextual information, the understanding of which is crucial for comprehensive analysis and effective modeling.

The rise of Large Language Models (LLMs), such as GPT~\cite{achiam2023gpt}, LLaMA~\cite{touvron2023llama}, and Gemini~\cite{team2023gemini}, has significantly advanced natural language processing. 
These multi-billion parameter models, pre-trained on extensive text corpora, have demonstrated impressive performance in natural language tasks, such as translation~\cite{zhang2023prompting,wang2023document}, question answering~\cite{lievin2024can,shi2023replug,kamalloo2023evaluating} and dialogue generation~\cite{zheng2023lmsys,qin2023chatgpt}.
Their remarkable few-shot and zero-shot learning capabilities allow them to understand diverse domains without requiring task-specific retraining or fine-tuning~\cite{brown2020language,yang2024harnessing,chang2023llm4ts}. 
Furthermore, they exhibit sophisticated reasoning and pattern recognition capabilities~\cite{mirchandani2023large,wang2023enhancing,chu2023leveraging}, enhancing their utility across various domains, including computer vision~\cite{koh2023grounding,guo2023images,pan2023retrieving,tsimpoukelli2021multimodal}, tabular data analysis~\cite{hegselmann2023tabllm,narayan2022can}, and audio processing~\cite{fathullah2024prompting,deshmukh2024training,tang2024extending}.

Motivated by the impressive general knowledge and reasoning abilities of LLMs, recent research has explored leveraging their strengths for time series (event) prediction.
For instance, pre-trained LLMs have been fine-tuned using time series data for specific tasks~\cite{zhou2024one,chang2023llm4ts}. 
Some studies introduce prompt tuning, where time series data is parameterized for input into either frozen LLMs~\cite{jin2023time,sun2023test} or fine-tunable LLMs~\cite{cao2023tempo}.
However, these methods typically use raw time series data or their parameterized embeddings, which are inherently distinct from the textual data that LLMs were pre-trained on, making it challenging for LLMs to utilize their rich semantic knowledge and contextual understanding capabilities.
One approach to address this limitation is prompting LLMs with textualized time series data, supplemented with simple contextual information, in a zero-shot manner~\cite{xue2023promptcast,liu2023large}.
However, the effectiveness of this approach is limited by the overly simplified contextualization of time series data.

\begin{figure*}[t]
    \centering
    \includegraphics[width=0.91\linewidth]{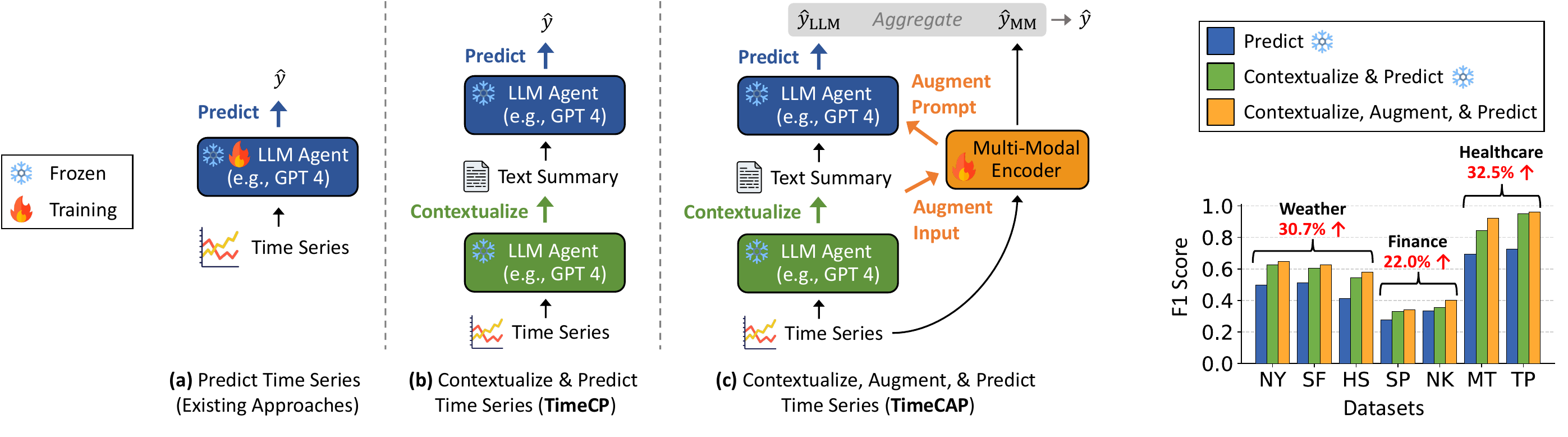}
    \caption{
    Approaches for time series event prediction using LLMs:
    \textbf{(a)} Existing methods use LLMs directly as predictors for time series data. 
    \textbf{(b)} Our \naive employs two LLM agents: the first agent, $\mathcal{A}_\text{C}$, contextualizes time series data into a text summary, and the second agent, $\mathcal{A}_\text{P}$, makes predictions based on this summary.
    \textbf{(c)} Our \method incorporates a multi-modal encoder that synergizes with LLM agents. 
    The multi-modal encoder generates predictions using both the generated text and the time series data. 
    Additionally, it samples relevant text from the training set to augment the prompt for $\mathcal{A}_\text{P}$ to make predictions. 
    \method achieves a 21.98\% improvement in F1 scores using contextualization alone and a 28.75\% improvement with the addition of augmentation for time series event predictions on real-world datasets.
    \label{fig:categorization}
    }
\end{figure*}

Notably, prior approaches leveraging LLMs for making predictions based on time series data have primarily focused on using LLMs as \textit{predictors}.
These methods either fine-tune LLMs or employ soft or hard prompting techniques, as illustrated in Figure~\ref{fig:categorization} (a). 
This approach often overlooks the importance of contextual understanding in time series analysis, such as the impact of geographical or climatic influences in weather prediction or the interdependencies between economic indicators in financial prediction. 
By harnessing the domain knowledge and contextual understanding capabilities of LLMs, we can uncover potential insights that can be overlooked by specialized time series models, leading to more comprehensive and accurate predictions.

In light of these insights, we present a novel framework that leverages LLMs not only as predictors but also as a \textit{contextualizer} of time series data.
Our preliminary method, \naive (\underline{\smash{C}}ontextualize \& \underline{\smash{P}}redict), incorporates two independent LLM agents, as illustrated in Figure~\ref{fig:categorization} (b).
The first agent generates a textual summary that provides a comprehensive contextual understanding of the input time series data, drawing on the LLM's extensive domain knowledge.
This summary is then used by the second agent to make more informed predictions of future events.
By contextualizing the time series data, \naive significantly enhances predictive performance compared to directly prompting LLMs with raw time series data or its parameterized embeddings.

Building upon \naive, we introduce \method (\underline{\smash{C}}ontextualize, \underline{\smash{A}}ugment, \& \underline{\smash{P}}redict), an advanced framework that further leverages text summaries generated by the first LLM agent as \textit{augmentations} to the time series data. 
\method incorporates a multi-modal encoder trained to predict events and learn representations using both the textual summaries and the raw time series data, as illustrated in Figure~\ref{fig:categorization} (c).
The representations learned by the multi-modal encoder are then used to select relevant text summaries from the training set, which are provided as in-context examples to augment the prompt for the second LLM agent.
This mutual enhancement, wherein the first LLM agent provides the encoder with contextualized information (i.e., input augmentation) and the enriched encoder supplies in-context examples to the second LLM agent (i.e., prompt augmentation), significantly improves overall performance.

Furthermore, \method is compatible with LLMs accessible through \textit{Language Modeling as a Service} (LMaaS)~\cite{sun2022black}, ensuring broader applicability to black-box APIs~\cite{achiam2023gpt}. 
Additionally, \method provides interpretable rationales for its predictions, addressing the critical need for transparency which is often overlooked in recent time series methods involving LLMs.

Our contributions are summarized as follows:
\begin{itemize}[leftmargin=*]
    \item \textbf{Novel framework.} We introduce \method, a novel and interpretable framework that leverages two LLM agents (as a contextualizer and a predictor) for event prediction using time series. It is further enhanced by mutual enhancement with a multi-modal encoder.
    \item \textbf{Prediction accuracy.} Our experimental results demonstrate that \method outperforms state-of-the-art methods in event prediction by up to 157\% in terms of F1 score.
    \item \textbf{Data contribution.} We collect seven real-world time series datasets from three different domains where underlying contextual understanding is crucial for effective time series modeling. 
    We release these datasets, along with the generated contextual text summaries by the LLM, to support future research and development in this domain.
\end{itemize}

\noindent \textbf{Code \& Datasets.} The {datasets} used in this paper are available at \url{https://github.com/geon0325/TimeCAP}. 
The {code} is available upon request.

\section{Related Work}
\label{sec:related}
\noindent\textbf{Large Language Models.}
In recent years, language models (LMs) such as BERT~\cite{devlin2019bert}, RoBERTa~\cite{liu2019roberta}, and DistilBERT~\cite{sanh2019distilbert} have evolved to large language models (LLMs) with multi-billion parameter architectures.~\footnote{We distinguish LMs (e.g., BERT), which are smaller and fine-tunable with academic resources, from LLMs (e.g., GPT-4), which are larger and generally infeasible to fine-tune in academic settings.}
LLMs, including GPT-4~\cite{achiam2023gpt}, LLaMA-2~\cite{touvron2023llama}, and PaLM~\cite{anil2023palm}, are trained on massive text corpora and have demonstrated impressive performance in various natural language tasks such as translation, summarization, and question answering.
These models possess extensive domain knowledge and exhibit zero-shot generalization capability, enabling them to perform tasks without specific training on those tasks~\cite{yang2024harnessing,brown2020language,kojima2022large}. 
Additionally, they exhibit emergent abilities such as arithmetic, multi-step reasoning, and instruction following, which LLMs were not explicitly trained for~\cite{wei2022emergent}. 
Their performance can be further enhanced through in-context learning, where a few input-label pairs are provided as demonstrations~\cite{brown2020language,min2022rethinking,liu2021makes}.
Their versatility has enabled adoption across various fields, including computer vision~\cite{koh2023grounding,guo2023images,pan2023retrieving,tsimpoukelli2021multimodal}, tabular data analysis~\cite{hegselmann2023tabllm,narayan2022can}, and audio processing~\cite{deshmukh2024training,tang2024extending}.

\noindent\textbf{LLMs and Time Series.}
Recent advancements in LLMs have attracted attention to their integration into time series analysis.
Approaches include training LLMs (or LMs) from scratch~\cite{ansari2024chronos,nie2022time,zhang2022crossformer} or fine-tuning pre-trained LLMs~\cite{zhou2024one,chang2023llm4ts}, using time series data.
Another approach is prompt tuning, where time series data is parameterized and input into either frozen LLMs~\cite{jin2023time,sun2023test} or trainable LLMs~\cite{cao2023tempo}.
These approaches bridge the gap between time series and LLMs by either integrating LLMs directly with time series data (LLM-for-time series) or aligning time series data with the LLM embedding spaces (time series-for-LLM)~\cite{sun2023test}.
Some studies use pre-trained LLMs without additional training (i.e., zero-shot prompting)~\cite{gruver2024large,liu2023large,xue2023promptcast}. 
For example, PromptCast~\cite{xue2023promptcast} textualizes time series inputs into prompts with basic contextual information.
For a more comprehensive overview, refer to recent surveys~\cite{jin2023large,jiang2024empowering,zhang2024large}.

\noindent\textit{\textbf{Our Work.}} Existing methods have focused on leveraging LLMs as direct predictors using time series through (fine-) tuning or soft/hard prompting. 
In this work, we utilize LLMs for two additional purposes beyond their typical role as a predictor.
Specifically, LLMs in \method play a role as a \textit{contextualizer} of time series data, providing a high-quality \textit{augmentation} that further enhances prediction performance.

\section{Proposed Method}
\label{sec:method}
We present our framework for predicting events based on time series using LLMs.
We begin with the problem statement.
Next, we present \naive, our initial method, which utilizes two LLM agents with distinct roles. 
Then, we describe \method, our ultimate version. 
Lastly, we discuss how \method offers interpretability for its predictions.

\subsection{Problem Statement}
We formally introduce LLMs and discuss the problem of time series event prediction.

\noindent\textbf{Large Language Models.}
Let us define an LLM $\mathcal{M}_\theta$, parameterized by $\theta$, which is pre-trained on extensive text corpora.
We keep $\theta$ fixed (i.e., frozen), employing LLMs in a zero-shot manner without any parameter updates or gradient computations, making them LMaaS-compatible. 
The LLM takes data of interest $D$ and optional supplementary data $S$ (e.g., demonstrations) to enhance understanding of $D$ and generate a more effective response $R$.
Utilizing a prompt generation function $p$, a prompt $p(D,S)$ is constructed, e.g., ``\textit{Refer to} $S$ and \textit{predict/summarize} $D$."
The inference of the LLM can thus be expressed as $R=\mathcal{M}_\theta (p(D, S))$.

In this context, we refer to \textit{LLM agents} as specialized instances of $\mathcal{M}_\theta$ designed to perform specific tasks.
Each LLM agent is tailored to leverage its pre-trained domain knowledge to address different aspects of time series event prediction. 
Their roles are determined by distinct prompt functions, such as predicting or summarizing the given data.

\noindent\textbf{Time Series Event Prediction.}
Given a time series $\bm{x}=(x_1,\cdots,x_L)$, where $L$ is the number of past timesteps and each $x_t\in \mathbb{R}^C$ represents data from $C$ channels at timestep $t$, the goal of time series event prediction is to predict the outcome $\bm{y}$ of a future event.
Real-world time series data (e.g., hourly humidity and temperature) is often associated with contextual information (e.g., geographical or climate factors) derived from domain knowledge.
This contextual information is crucial for accurate future event predictions (e.g., forecasting next-day rain).
We define the problem as a multi-class classification task and leave the exploration of regression-based event forecasting for future work.

\subsection{\naive: Contextualize and Predict}
We introduce \naive, our preliminary method for LLM-based time series event prediction.
\naive leverages the contextual information associated with time series data to enhance the comprehension and predictive capabilities of LLMs in a zero-shot manner.

PromptCast~\cite{xue2023promptcast} is a direct counterpart to \naive, as it prompts the LLM with a textualized prompt of time series to make predictions.
However, it focuses on using LLMs as a predictor and does not fully utilize their contextualization capabilities.

To address this limitation, \naive introduces two independent LLM agents, $\mathcal{A}_\text{C}$ and $\mathcal{A}_\text{P}$, which aim to better leverage the contextualization capabilities of LLMs for time series event prediction.
The first agent, $\mathcal{A}_\text{C}$, generates a textual summary $\bm{s}_{\bm{x}}$ that contains the underlying context of the given time series $\bm{x}$ by leveraging its domain knowledge:
\begin{equation*}
 \bm{s}_{\bm{x}} = \mathcal{A}_\text{C}(\bm{x})=\mathcal{M}_\theta(p_\text{C}(\bm{x})),
\end{equation*}
where $p_\text{C}(\bm{x})$ is a prompt that instructs the LLM to \textit{contextualize} $\bm{x}$.
The generated summary, $\bm{s}_{\bm{x}}$, includes relevant contextual insights beyond the raw time series data $\bm{x}$, which is then used by the second agent, $\mathcal{A}_{\text{P}}$, to make more informed event predictions:
\begin{equation*}
    \hat{\bm{y}}_{\text{LLM}} = \mathcal{A}_\text{P}(\bm{s}_{\bm{x}}) = \mathcal{M}_\theta(p_\text{P}(\bm{s}_{\bm{x}})),
\end{equation*}
where $p_\text{P}(\bm{s}_{\bm{x}})$ is a prompt that instructs the LLM to \textit{predict} the outcome of the event based on $\bm{s}_{\bm{x}}$.
By incorporating the context-informed summary generated by $\mathcal{A}_\text{C}$, $\mathcal{A}_\text{P}$ can account for the broader context in its predictions.
As shown in Figure~\ref{fig:categorization}, our dual-agent-based approach consistently outperforms the single-agent approach (spec., PromptCast), which directly predicts the event from the input time series data, i.e., $\mathcal{M}_\theta(p_\text{P}(\bm{x}))$. 
The enhanced accuracy demonstrates the effectiveness of generating and utilizing contextual information for future event predictions with LLMs.

\subsection{\method: Contextualize, Augment, Predict}
Building upon \naive, we present \method, an advanced version of our framework. 
\method trains a multi-modal encoder that synergizes with the LLM agents by introducing dual augmentations (spec., input and prompt augmentations) where the multi-modal encoder and LLM agents complement each other, enabling \method to make more accurate and reliable event predictions.

\begin{figure}
    \centering
    \includegraphics[width=1.00\columnwidth]{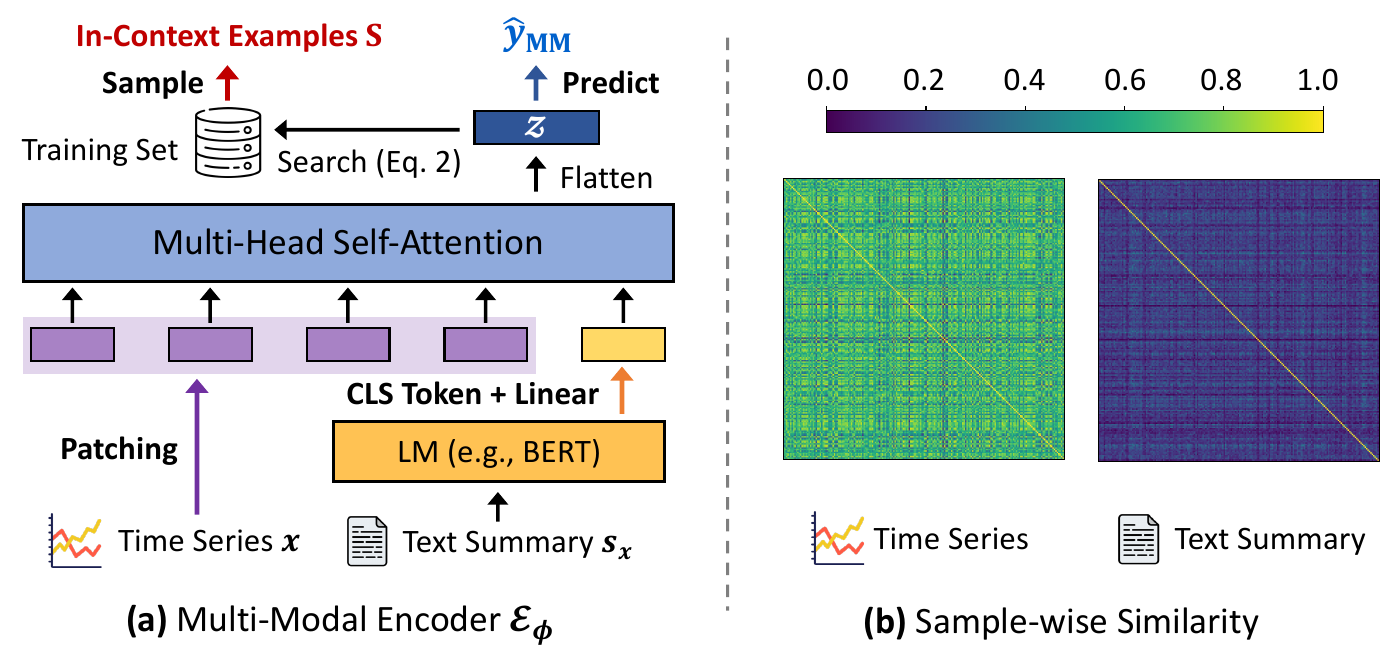}
    \caption{
    \textbf{(a)} The multi-modal encoder $\mathcal{E}_\phi$ generates an embedding $\bm{z}$ and a prediction $\hat{\bm{y}}_\text{MM}$ based on the multi-modal input $(\bm{x}, \bm{s}_{\bm{x}})$, i.e., time series and its augmented text summary (Eq.~\eqref{eq:mm}). The generated embedding $\bm{z}$ is used to retrieve relevant summaries from the training set to serve as in-context examples to augment the prompt for $\mathcal{A}_\text{P}$ (Eq.~\eqref{eq:nearest}).
    \textbf{(b)} The similarity patterns within the time series and the text vary; time series similarities are generally high, while text similarities are selectively highlighted, implying complementary information in each modality.\protect\footnotemark}
    \label{fig:encoder}
\end{figure}
\footnotetext{The similarity between time series is computed using negative Dynamic Time Warping~\cite{berndt1994using}, and the similarity between texts is computed using TF-IDF~\cite{chowdhury2010introduction}. 
}

\noindent\textbf{Multi-Modal Encoder.}
We introduce a trainable encoder $\mathcal{E}_\phi$, parameterized by $\phi$ (Figure~\ref{fig:encoder} (a)).
This encoder aims to capture intricate dynamic patterns in time series data more effectively than zero-shot LLMs.
In addition to time series $\bm{x}$, it incorporates the textual summary $\bm{s}_{\bm{x}}$ generated by $\mathcal{A}_\text{C}$, which provides additional contextual insights beyond the raw time series data (i.e., \textit{input augmentation}), as shown in Figure~\ref{fig:encoder} (b). 
The encoder $\mathcal{E}_\phi$ generates its prediction $\hat{\bm{y}}_{\text{MM}}$ and the embedding $\bm{z}$ of the multi-modal input $(\bm{x},\bm{s}_{\bm{x}})$ as:
\begin{equation}\label{eq:mm}
    (\hat{\bm{y}}_{\text{MM}}, \;\bm{z}) = \mathcal{E}_\phi(\bm{x}, \bm{s}_{\bm{x}}),
\end{equation}
where $\bm{z}$ is used for sampling in-context examples (Eq.~\eqref{eq:nearest}).
The encoder consists of (1) a language model that embeds text into the latent space and (2) a transformer encoder that captures dependencies between the two modalities.

The corresponding text summary $\bm{s}_{\bm{x}}$ is processed by a pre-trained language model (LM) with substantially fewer parameters, which is thus relatively easier to fine-tune.
Specifically, we represent $\bm{s}_{\bm{x}}$ using the LM as $\hat{\bm{z}}=\mathtt{LM}(\bm{s}_{\bm{x}})\in \mathbb{R}^{d'}$, leveraging the output CLS token embedding.
This representation is then projected as $\tilde{\bm{z}}_{\text{text}}=\hat{\bm{z}}\mathbf{W}_{\text{text}}\in \mathbb{R}^{d}$ using a linear layer $\mathbf{W}_{\text{text}}\in\mathbb{R}^{d'\times d}$.

For time series, motivated by the effectiveness of patching~\cite{nie2022time,zhang2022crossformer}, we segment a time series $\bm{x}^{(i)}\in\bm{x}$ of the $i$\textsuperscript{th} channel into $N$ non-overlapping patches $\hat{\bm{x}}^{(i)}\in \mathbb{R}^{N\times L_p}$ with patch length $L_p$ and stride $L_s$, where $N= \lceil \frac{L-L_p}{L_s} \rceil+1$ holds.
These patches are then projected as $\tilde{\bm{z}}_{\text{time}}^{(i)} = \hat{\bm{x}}^{(i)}\mathbf{W}_{\text{time}} \in \mathbb{R}^{N\times d}$ using a simple linear layer $\mathbf{W}_{\text{time}}\in\mathbb{R}^{L_p\times d}$.

For each $i$\textsuperscript{th} channel of the time series, we concatenate the time series patch embeddings $\tilde{\bm{z}}_{\text{time}}^{(i)}$ and the text embedding $\tilde{\bm{z}}_{\text{text}}$ to construct $\tilde{\bm{z}}^{(i)}=[\tilde{\bm{z}}_{\text{time}}^{(i)}; \tilde{\bm{z}}_{\text{text}}]\in \mathbb{R}^{(N+1)\times d}$.
We then use multi-head self-attention to capture the relationships between this combined representation.
More specifically, for each attention head $h\in \{1,\cdots,H\}$, we compute query $\mathbf{Q}_h^{(i)}=\tilde{\bm{z}}^{(i)}\mathbf{W}_h^Q$, key $\mathbf{K}_h^{(i)}=\tilde{\bm{z}}^{(i)}\mathbf{W}_h^K$, and value $\mathbf{V}_h^{(i)}=\tilde{\bm{z}}^{(i)}\mathbf{W}_h^V$ matrices using $\mathbf{W}_h^Q,\mathbf{W}_h^K,\mathbf{W}_h^V\in\mathbb{R}^{d\times d/H}$.
Each $h$\textsuperscript{th} attention head is then defined as:
\begin{equation*}
    \scalebox{0.99}{$
    \bm{z}_h^{(i)} = \mathtt{Softmax}\left( \frac{\mathbf{Q}_h^{(i)} {\mathbf{K}_h^{(i)}}^T }{\sqrt{d/H}} \right) \mathbf{V}_h^{(i)}.
    $}
\end{equation*}
The outputs of the attention heads are aggregated and projected as $\bm{z}^{(i)} = [\bm{z}_1^{(i)} ; \cdots ; \bm{z}_H^{(i)} ] \mathbf{W}^H \in \mathbb{R}^d$ where $\mathbf{W}^H\in\mathbb{R}^{d\times d}$.
Then, a flatten layer represents all channels as a single embedding vector, i.e., $\bm{z}=[\bm{z}^{(1)};\cdots; \bm{z}^{(C)}] \in \mathbb{R}^{dC}$.
Finally, a linear layer $\mathbf{W}^P\in\mathbb{R}^{dC\times K}$ is applied to $\bm{z}$ to obtain a $K$-dimensional prediction logit, i.e,. $\hat{\bm{y}}_{\text{MM}}=\bm{z}\mathbf{W}^P\in \mathbb{R}^K$.
We fine-tune the parameters $\phi$ including those of the LM and the transformer encoder using cross-entropy loss.

\begin{table*}[h]
\centering
\scalebox{0.8}{
\begin{tabular}{c@{\hspace{4pt}}c|c@{\hspace{4pt}}c@{\hspace{4pt}}c@{\hspace{4pt}}c|c|l}
    \toprule
     Domain & Dataset & Resolution & \# Channels & \# Timestamps & \# Samples & Duration & Label Distribution\\
    \midrule
    \multirow{3}{*}{Weather} & New York (NY) & Hourly & 5 & 45,216 & 1,884 & 2012.10 - 2017.11 & Rain (24.26\%) / Not rain (75.74\%)\\
    & San Francisco (SF) & Hourly & 5 & 45,216 & 1,884 & 2012.10 - 2017.11 & Rain (24.58\%) / Not rain (75.42\%) \\
    & Houston (HS) & Hourly & 5 & 45,216 & 1,884 & 2012.10 - 2017.11 & Rain (30.94\%) / Not rain (69.06\%)\\
    \midrule
    \multirow{2}{*}{Finance} & S\&P 500 (SP) & Daily & 9 & 1,258 & 1,238 & 2019.01 - 2023.12 & Inc. (13.78\%) / Dec. (17.04\%) / Etc. (69.18\%)\\
    & Nikkei 225 (NK) & Daily & 9 & 1,258 & 1,238 & 2019.01 - 2023.12 & Inc. (15.02\%) / Dec. (17.12\%) / Etc. (67.86\%)\\
    \midrule
    \multirow{2}{*}{Healthcare} & Mortality (MT) & Weekly & 4 & 395 & 375 & 2016.07 - 2024.06  & Exceed (69.33\%) / Not exceed (30.67\%) \\
    & Test-Positive (TP) & Weekly & 6 & 447 & 427 & 2015.10 - 2024.04 & Exceed (65.77\%) / Not exceed (34.23\%)  \\
    \bottomrule
\end{tabular}}
\caption{\label{tab:dataset} 
Statistics of seven real-world datasets for time series event prediction. These publicly available datasets are expected to benefit from contextual understanding beyond raw time series data. More details can be found in the supplementary document.
}
\end{table*}

\noindent\textbf{In-Context Example Sampling.}
Once the multi-modal encoder is trained, it aids $\mathcal{A}_{\text{P}}$ in making more informed predictions by sampling relevant text summaries from the training set as valuable demonstrations (i.e., \textit{prompt augmentation}). 

Given the embedding $\bm{z}$ of the multi-modal input ($\bm{x}$, $\bm{s}_{\bm{x}}$), we retrieve $k$ summaries from the training set whose embeddings are closest to $\bm{z}$. 
Formally, let $\mathcal{T}$ denote the training set, and $\bm{\mathcal{Z}}\in \mathbb{R}^{|\mathcal{T}|\times dC}$ represent the set of embeddings of the training samples generated by $\mathcal{E}_\phi$.
The $k$ pairs of text summaries and their corresponding outcomes are retrieved as the nearest neighbors of $\bm{z}$ in the embedding space, as follows:
\begin{align}\label{eq:nearest}
    &\mathbf{S} =\{(\bm{s}_{\bm{x}_j}, \bm{y}_j) : j \in \text{NN}_k(\bm{z})\}, \\
    \text{where}\;\;\;\;& \text{NN}_k(\bm{z}) = {\text{arg}\;\text{top-}{k}}_{j\in \mathcal{T}}(-\|\bm{z} - \bm{\mathcal{Z}}_j\|). \notag
\end{align}
These summaries and their outcomes are used as in-context examples for $\mathcal{A}_\text{P}$ to predict the outcome of $\bm{s}_{\bm{x}}$ as follows:
\begin{equation}\label{eq:lmm}
    \hat{\bm{y}}_{\text{LLM}} =\mathcal{A}_\text{P} \left(\bm{s}_{\bm{x}}, \mathbf{S}\right) = \mathcal{M}_\theta \left(p_\text{P}(\bm{s}_{\bm{x}}, \mathbf{S})\right).
\end{equation}
These examples help the agent $\mathcal{A}_\text{P}$ better understand the test input by comparing the summaries and reasoning based on them. 
This leads to more accurate predictions, as shown in Figure~\ref{fig:categorization} and further validated in Section~\ref{sec:exp}.

\noindent\textbf{Fused Prediction.}
Lastly, we integrate the predictions from the multi-modal encoder (Eq.~\eqref{eq:mm}) and $\mathcal{A}_\text{P}$ (Eq.~\eqref{eq:lmm}) through a linear combination, i.e., $\hat{\bm{y}}=\lambda \hat{\bm{y}}_{\text{LLM}} + (1 - \lambda) \hat{\bm{y}}_{\text{MM}}$ where $\lambda\in [0,1]$ is a hyperparameter.
Given that the prediction $\hat{\bm{y}}_{\text{LMM}}$ produced by $\mathcal{A}_\text{P}$ is discrete, we convert it into a one-hot vector to enable its fusion with the continuous logit $\hat{\bm{y}}_{\text{MM}}$.
This fusion leverages complementary information from both models, enhancing the overall performance of \method, as demonstrated in Section~\ref{sec:exp}.

\subsection{Interpreting the Predictions\label{sec:method:interpretation}}
We explore how \method provides interpretations for its predictions by introducing two variants of the prompt function $p_\text{P}$ used in $\mathcal{A}_\text{P}$.
The resulting variants, $\mathcal{A}_{\text{P}}^{\text{I}}$ and $\mathcal{A}_{\text{P}}^{\text{E}}$, enable distinct interpretations, enhancing transparency.

\noindent\textbf{Implicit Interpretation.}
We prompt the LLM $\mathcal{M}_\theta$ to generate both a prediction and its corresponding rationale:
\begin{equation*}
    (\hat{\bm{y}}_{\text{LLM}},\; \bm{r})  = \mathcal{A}_\text{P}^{\text{I}}(\bm{s}_{\bm{x}}, \mathbf{S}) = \mathcal{M}_\theta\left( p_{\text{P}}^{\text{I}}(\bm{s}_{\bm{x}}, \mathbf{S}) \right),
\end{equation*}
where $p_{\text{P}}^{\text{I}}$ is a prompt function that instructs $\mathcal{M}_\theta$ to predict the event ($\hat{\bm{y}}_{\text{LLM}}$) and also provide the rationale ($\bm{r}$) behind its prediction.
This rationale leverages the LLM's domain knowledge and reasoning capabilities.
While the in-context examples $\mathbf{S}$ are optional, as shown in Section~\ref{sec:exp}, their inclusion leads to distinct implicit interpretations.

\noindent\textbf{Explicit Interpretation.} 
We prompt $\mathcal{M}_\theta$ to identify the most useful or relevant example from the in-context set $\mathbf{S}$:
\begin{equation*}
    (\hat{\bm{y}}_{\text{LLM}},\; \bm{s}_{\bm{x}_{j^*}} ) =\mathcal{A}_\text{P}^{\text{E}}(\bm{s}_{\bm{x}}, \mathbf{S}) = \mathcal{M}_\theta\left( p_{\text{P}}^{\text{E}}(\bm{s}_{\bm{x}}, \mathbf{S}) \right),
\end{equation*}
where $p_{\text{P}}^{\text{E}}$ is a prompt function that instructs $\mathcal{M}_\theta$ to predict the event ($\hat{\bm{y}}_{\text{LLM}}$) and select the most relevant example ($\bm{s}_{\bm{x}_{j^*}}$) from $\mathbf{S}$.
In addition, the input time series $\bm{x}$ can be compared with the corresponding time series $\bm{x}_{j^*}$ for further analyses.

\section{Experiments}
\label{sec:exp}
In this section, we present \method's: (1) accuracy compared with the state-of-the-art methods, (2) component effectiveness, (3) interpretability, and (4) additional analyses.~\footnote{The code is available upon request. }

\subsection{Experimental Settings\label{sec:exp:settings}}
We first report the experimental settings.
We use GPT-4~\cite{achiam2023gpt} as the default backbone for the LLM agents and BERT~\cite{devlin2019bert} as the LM within the multi-modal encoder. 
We describe the prompt functions in the supplementary document. 

\begin{table*}[t!]
\centering
\scalebox{0.76}{
\begin{tabular}{l@{\hspace{4.1pt}}|c@{\hspace{4.1pt}}c|c@{\hspace{4.1pt}}c|c@{\hspace{4.1pt}}c|c@{\hspace{4.1pt}}c|c@{\hspace{4.1pt}}c|c@{\hspace{4.1pt}}c|c@{\hspace{4.1pt}}c|c@{\hspace{4.1pt}}c}
    \toprule
    & \multicolumn{6}{c|}{\textbf{Weather}} & \multicolumn{4}{c|}{\textbf{Finance}} & \multicolumn{4}{c|}{\textbf{Healthcare}} & & \\
    \textbf{Datasets $\rightarrow$} & \multicolumn{2}{c|}{\textbf{New York}} & \multicolumn{2}{c|}{\textbf{San Fran.}} & \multicolumn{2}{c|}{\textbf{Houston}} & \multicolumn{2}{c|}{\textbf{S\&P 500}} & \multicolumn{2}{c|}{\textbf{Nikkei 225}} & \multicolumn{2}{c|}{\textbf{Mortality}} & \multicolumn{2}{c|}{\textbf{Test-Positive}} & \multicolumn{2}{c}{\textbf{Avg. Rank}}\\
    \textbf{Methods $\downarrow$} & \textbf{F1} & \textbf{AUC} & \textbf{F1} & \textbf{AUC} & \textbf{F1} & \textbf{AUC} & \textbf{F1} & \textbf{AUC} & \textbf{F1} & \textbf{AUC} & \textbf{F1} & \textbf{AUC} & \textbf{F1} & \textbf{AUC} & \textbf{F1} & \textbf{AUC}\\
    \midrule
    Autoformer~\cite{wu2021autoformer} & 0.546 & 0.590 & 0.475 & 0.539 & 0.542 & 0.592 & 0.330 & 0.471 & 0.358 & 0.568 & 0.683 & 0.825 & 0.774 & 0.918 & 9.14 & 10.00\\
    Crossformer~\cite{zhang2022crossformer} & 0.500 & 0.594 & 0.546 & 0.594 & 0.611 & 0.672 & 0.330 & 0.561 & 0.283 & 0.610 & 0.737 & 0.914 & 0.924 & \underline{\smash{0.984}} & 6.86 & 5.14 \\
    TimesNet~\cite{wu2022timesnet} & 0.494 & 0.594 & 0.521 & 0.557 & \textbf{0.614} & 0.663 & 0.288 & 0.566 & 0.272 & 0.473 & 0.558 & 0.903 & 0.794 & 0.867 & 9.57 & 8.57 \\
    DLinear~\cite{zeng2023transformers} & 0.540 & 0.660 & 0.553 & \underline{\smash{0.633}} & 0.592 & 0.669 & 0.174 & 0.463 & 0.278 & 0.509 & 0.419 & 0.388 & 0.393 & 0.500 & 10.43 & 9.43 \\
    TSMixer~\cite{chentsmixer} & 0.488 & 0.534 & 0.577 & 0.577 & 0.522 & 0.589 & \textbf{0.405} & \underline{\smash{0.567}} & 0.367 & 0.516 & 0.808 & 0.931 & 0.550 & 0.600 & 7.43 & 9.00 \\
    PatchTST~\cite{nie2022time} & 0.592 & 0.675 & 0.542 & 0.565 & 0.593 & 0.652 & 0.373 & \textbf{0.573} & \underline{\smash{0.391}} & \textbf{0.640} & 0.695 & 0.928 & 0.841 & 0.934 & 5.14 & \underline{\smash{5.00}}  \\
    FreTS~\cite{yi2024frequency} & \underline{\smash{0.625}} & 0.689 & 0.504 & 0.533 & 0.592 & 0.673 & 0.351 & 0.532 & 0.381 & 0.575 & 0.464 & 0.500 & 0.817 & 0.812 & 6.86 & 8.00 \\
    iTransformer~\cite{liuitransformer} & 0.541 & 0.650 & 0.534 & 0.566 & 0.569 & 0.655 & 0.285 & 0.537 & 0.269 & 0.462 & 0.797 & 0.972 & 0.887 & 0.950 & 8.71 & 7.29 \\
    \midrule
    LLMTime~\cite{kojima2022large}~\textcolor{blue}{\textsuperscript{\ding{91}}} & 0.587 & 0.657 & 0.542 & 0.563 & 0.587 & 0.626 & 0.306 & 0.492 & 0.166 & 0.510 & 0.769 & 0.804 & 0.802 & 0.817 & 8.43 & 9.86 \\
    PromptCast~\cite{xue2023promptcast}~\textcolor{blue}{\textsuperscript{\ding{91}}} & 0.499 & 0.365 & 0.510 & 0.397 & 0.412 & 0.400 & 0.276 & 0.488 & 0.333 & 0.517 & 0.695 & 0.869 & 0.727 & 0.768 & 11.00 & 12.00 \\
    GPT4TS~\cite{zhou2024one} & 0.501 & 0.606 & 0.550 & 0.612 & 0.612 & \textbf{0.692} & 0.285 & 0.414 & 0.297 & 0.531 & \underline{\smash{0.901}} & \underline{\smash{0.992}} & 0.774 & 0.879 & 7.00 & 6.14 \\
    Time-LLM~\cite{jin2023time} & 0.613 & 0.699 & 0.577 & 0.593 & 0.592 & 0.625 & 0.357 & 0.552 & 0.294 & 0.526 & 0.659 & 0.926 & 0.671 & 0.864 & 7.00 & 6.86 \\
    \midrule
    \textbf{\naive}~\textcolor{blue}{\textsuperscript{\ding{91}}} & \underline{\smash{0.625}} & \underline{\smash{0.706}} & \underline{\smash{0.603}} & 0.607 & 0.544 & 0.593 & 0.330 & 0.510 & 0.364 & 0.532 & 0.842 & 0.946 & \underline{\smash{0.949}} & 0.956 & \underline{\smash{4.43}} & 5.57 \\
    \textbf{\method} & \textbf{0.676} & \textbf{0.745} & \textbf{0.632} & \textbf{0.676} & \textbf{0.614} & \underline{\smash{0.675}} & \underline{\smash{0.398}} & 0.546 & \textbf{0.428} & \textbf{0.640} & \textbf{0.947} & \textbf{1.000} & \textbf{0.962} & \textbf{0.995} & \textbf{1.14} & \textbf{1.86} \\
    \bottomrule
\end{tabular}}
\caption{\label{tab:result} The F1 score (F1) and the AUROC (AUC) for \naive, \method, and their competitors on seven real-world time series datasets. \method outperforms other methods on most datasets and ranks first on average. Methods annotated with \textcolor{blue}{\textsuperscript{\ding{91}}} make predictions in a zero-shot manner. The best and second-best scores are highlighted in \textbf{bold} and \underline{\smash{underline}}, respectively.}
\end{table*}

\noindent\textbf{Datasets and Tasks.}
We collected seven real-world time series datasets from three domains: weather, finance, and healthcare, for time series event prediction, as summarized in Table~\ref{tab:dataset}.
Note that only time series data is provided, and the text data is generated by $\mathcal{A}_\text{C}$. 
Below, we describe the datasets and their respective tasks in each domain.
\begin{itemize}[leftmargin=*]
    \item \textbf{Weather}: Datasets in this domain consist of hourly time series data on temperature, humidity, air pressure, wind speed, and wind direction in New York (\textbf{NY}), San Francisco (\textbf{SF}), and Houston (\textbf{HS}).
    Given the last 24 hours of time series data, the task is to predict the event of whether it will rain in the next 24 hours.
    \item \textbf{Finance}: Datasets in this domain contain daily time series data for nine financial indicators (e.g., S\&P 500, VIX, and exchange rates).
    The task is to predict whether the price of S\&P 500 (\textbf{SP}) or Nikkei 225 (\textbf{NK}) will (1) increase by more than 1\%, (2) decrease by more than 1\%, or (3) otherwise remain relatively stable.
    \item \textbf{Healthcare}: The mortality (\textbf{MT}) dataset includes weekly data including influenza and pneumonia deaths, and the task is to predict if the mortality ratio from influenza/pneumonia will exceed the average threshold. 
    The test-positive (\textbf{TP}) dataset includes weekly data including the number of positive specimens for Influenza A \& B, and the task is to predict if the ratio of respiratory specimens testing positive for influenza will exceed the average threshold.
\end{itemize}
We release the datasets with their text summaries generated by $\mathcal{A}_\text{C}$.~\footnote{\url{https://github.com/geon0325/TimeCAP}}
See the supplementary document for more details. 

\noindent\textbf{Baselines.}
We consider state-of-the-art time series prediction models, including Autoformer~\cite{wu2021autoformer}, Crossformer~\cite{zhang2022crossformer}, TimesNet~\cite{wu2022timesnet}, DLinear~\cite{zeng2023transformers}, PatchTST~\cite{nie2022time}, FreTS~\cite{yi2024frequency}, and iTransformer~\cite{liuitransformer} as well as recent LLM-based models, including LLMTime~\cite{kojima2022large}, PromptCast~\cite{xue2023promptcast}, GPT4TS~\cite{zhou2024one}, and Time-LLM~\cite{jin2023time}, as baselines.
While these methods are primarily designed for regression-based time series prediction, they can be easily adapted for event prediction (i.e., classification) tasks.~\footnote{\url{https://github.com/thuml/Time-Series-Library}}
See the supplementary document for more details.

\noindent\textbf{Evaluation.}
We evaluate \method and its competitors on time series event prediction using the F1 Score and AUROC.
We split data into training, validation, and test sets in a 6:2:2 ratio and set $k=5$, unless otherwise stated.
We run five times for each setting.
For other hyperparameter settings, refer to the supplementary document.

\subsection{Accuracy\label{sec:exp:accuracy}}
We first report the predictive performance of \method and its competitors on time series event prediction.

\noindent\textbf{Main Results.} 
Table~\ref{tab:result} presents the performance of \method, \naive, and their competitors across all datasets. 
\method achieves the best average performance and overall ranks. 
These results demonstrate the effectiveness of our framework in contextualizing time series data and the mutual enhancement with the multi-modal encoder.

\noindent\textbf{Zero-shot Results.}
\naive predicts events in a zero-shot manner without referencing other training samples. 
As shown in Table~\ref{tab:result}, \naive significantly outperforms other zero-shot LLM-based methods (LLMTime~\cite{kojima2022large} and PromptCast~\cite{xue2023promptcast}) across all datasets. 
While competitors use LLMs mainly as predictors, \naive leverages LLMs' domain knowledge and reasoning capabilities to understand the context of the time series, which leads to more accurate predictions.

\begin{table}[t]
  \centering
  \scalebox{0.685}{
    \begin{tabular}{l@{\hspace{2.9pt}}l|c@{\hspace{2.9pt}}c@{\hspace{2.9pt}}c|c@{\hspace{2.9pt}}c|c@{\hspace{2.9pt}}c|c}
        \toprule
        & & \multicolumn{3}{c|}{\textbf{Weather}} & \multicolumn{2}{c|}{\textbf{Finance}} & \multicolumn{2}{c|}{\textbf{Healthcare}} & \textbf{Avg.} \\
        & & \textbf{NY} & \textbf{SF} & \textbf{HS} & \textbf{SP} & \textbf{NK} & \textbf{MT} & \textbf{TP} & \textbf{Rank}\\
        \midrule
        \multirow{2}{*}{\makecell{(1) Context}} & PromptCast & 0.499 & 0.510 & 0.412 & 0.276 & 0.333 & 0.695 & 0.727  & 8.86 \\
        & \naive & 0.625 & 0.603 & 0.544 & 0.330 & 0.364 & 0.842 & 0.949 & 5.57 \\
        \midrule
        \multirow{2}{*}{\makecell{(2) Input}} & Only Time & 0.592 & 0.542 & 0.593 & 0.373 & 0.391 & 0.695 & 0.841   & 6.43\\
        & Time+Text & 0.623 & 0.576 & \underline{\smash{0.606}} & \textbf{0.398} & \textbf{0.428} & 0.734 & 0.937  & 4.29\\
        \cmidrule(lr){1-10}
        \multirow{3}{*}{\makecell{(3) Prompt}} & Random & 0.619 & 0.621 & 0.528 & 0.322 & 0.344 & 0.901 & 0.883  & 6.57 \\
        & Only Time & 0.625 & 0.599 & 0.571 & 0.305 & 0.379 & \textbf{0.947} & 0.948 & 5.14 \\
        &  Time+Text & \underline{\smash{0.641}} & \underline{\smash{0.626}} & 0.578 & 0.341 & 0.401 & \textbf{0.947} & \underline{\smash{0.961}} & 3.00 \\
        \midrule
        \multirow{2}{*}{\makecell{(4) Fusion}} & Select-One & \underline{\smash{0.641}} & \underline{\smash{0.626}} & \underline{\smash{0.606}} & \textbf{0.398} & \textbf{0.428} & \textbf{0.947} & \underline{\smash{0.961}}  & \underline{\smash{1.57}} \\
        & Aggregate & \textbf{0.676} & \textbf{0.632} & \textbf{0.614} & \textbf{0.398} & \textbf{0.428} & \textbf{0.947} & \textbf{0.962}  & \textbf{1.00} \\
        \bottomrule
    \end{tabular}}
    \caption{\label{tab:ablation} Ablation studies of \method. 
      Every component of \method contributes to the improvement of the predictive performance in terms of the F1 score. 
      }
\end{table}

\begin{figure*}[t]
    \centering
    \includegraphics[width=1.0\linewidth]{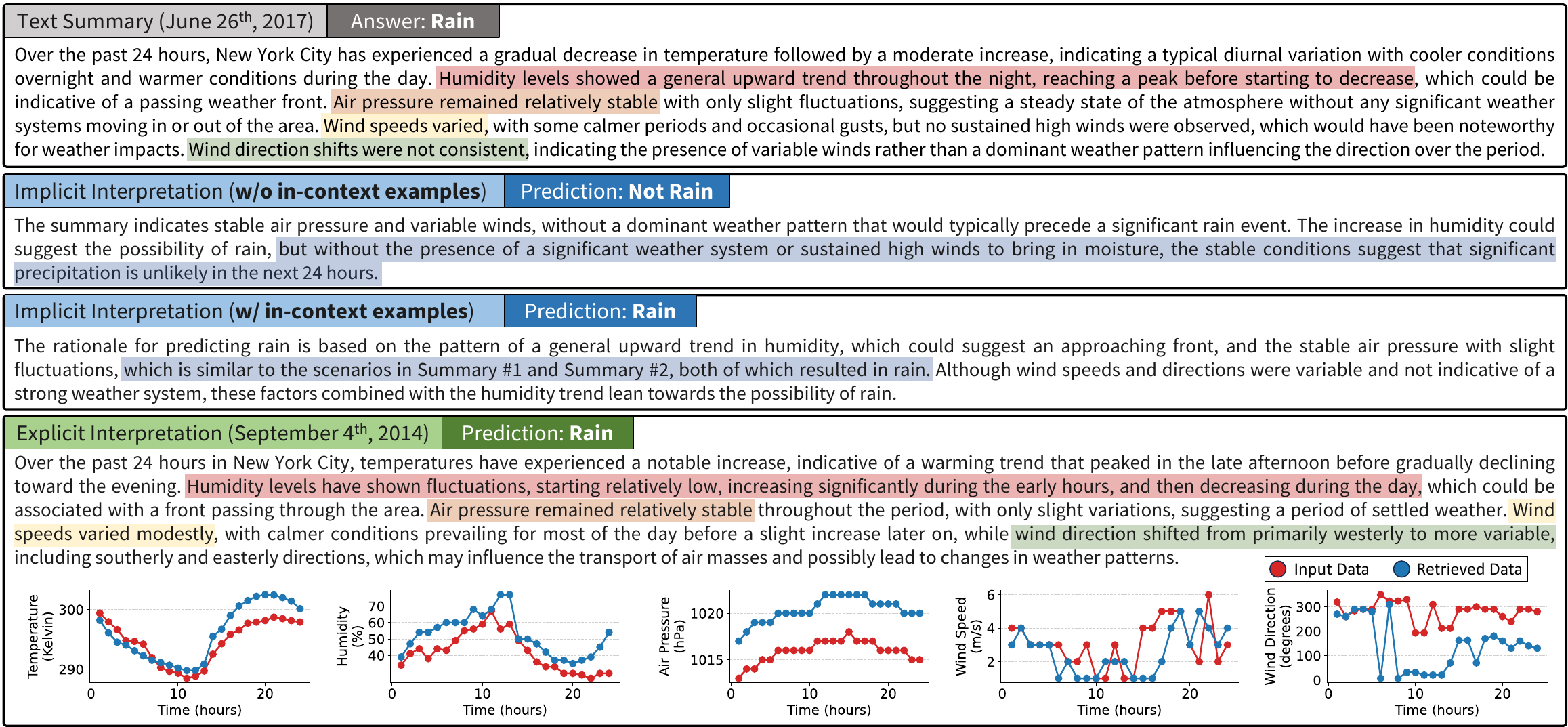}
    \caption{
    A case study on interpretations of \method. 
    Given a text summary: \textbf{(a)} implicit interpretations depend on the presence of in-context examples (\hllightblue{blue}), and
    \textbf{(b)} explicit interpretations involve post-hoc comparisons between the input and a selected in-context example with similar semantics (\hlpink{red}, \hlorange{orange}, \hlyellow{yellow}, and \hlgreen{green}).
    }
    \label{fig:interpretation}
\end{figure*}

\begin{figure}
    \centering
    \includegraphics[width=0.525\linewidth]{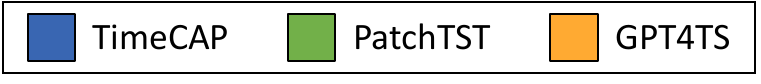}\\
    \begin{subfigure}[t]{.16\textwidth}
        \centering
        \includegraphics[width=0.99\linewidth]{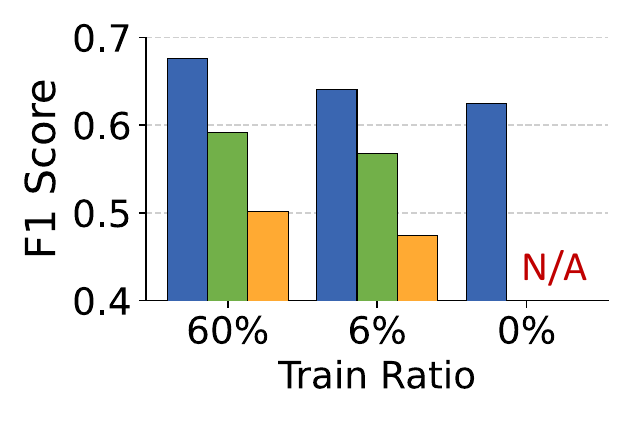}
        \captionsetup{justification=centering}
        \caption{Weather (NY)}
    \end{subfigure}
    \hspace{-7pt}
    \begin{subfigure}[t]{.16\textwidth}
        \centering
        \includegraphics[width=0.99\linewidth]{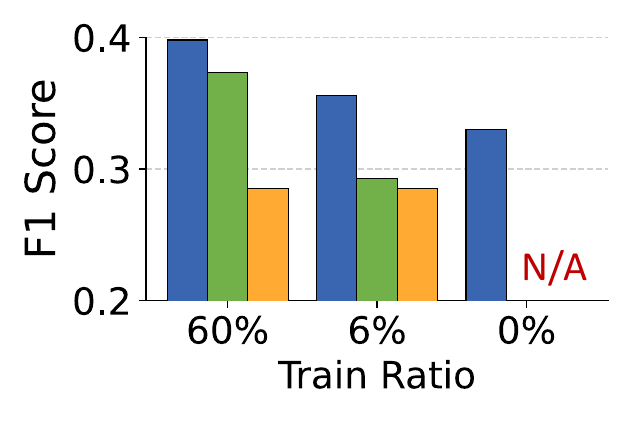}
        \captionsetup{justification=centering}
        \caption{Finance (SP)}
    \end{subfigure}
    \hspace{-7pt}
    \begin{subfigure}[t]{.16\textwidth}
        \centering
        \includegraphics[width=0.99\linewidth]{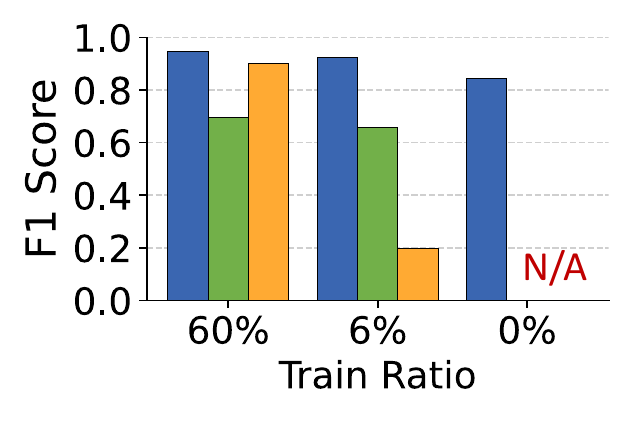}
        \captionsetup{justification=centering}
        \caption{Healthcare (MT)}
    \end{subfigure}
    \caption{\method consistently outperforms its competitors (spec., PatchTST and GPT4TS) across different training ratios. When the training ratio is 0\%, \naive is used.}
    \label{fig:few}
\end{figure}

\subsection{Effectiveness\label{sec:exp:effectiveness}}

We verify the effectiveness of each component of \method through ablation studies as summarized in Table~\ref{tab:ablation}. 

\noindent\textbf{Contextualization.}
As shown in (1) in Table~\ref{tab:ablation}, \naive consistently and significantly outperforms PromptCast, which directly prompts LLMs to predict the future event, across all datasets.
This demonstrates the effectiveness \naive's dual-agent approach in contextualizing time series data for event prediction.

\noindent\textbf{Augmentation.}
\method incorporates dual augmentations: $\mathcal{A}_\text{C}$ generates textual summaries to augment the time series data (i.e., input augmentation), and the multi-modal encoder samples in-context examples to augment prompts for $\mathcal{A}_\text{P}$ (i.e., prompt augmentation). 
We evaluate the effectiveness of each augmentation.

\begin{itemize}[leftmargin=*]
    \item \textbf{Input Augmentation.} 
    Our multi-modal encoder, which incorporates both time series and text data, consistently outperforms its variant that relies only on time series data, as shown in (2) of Table~\ref{tab:ablation}.
    This demonstrates that the textual summaries generated by $\mathcal{A}_\text{C}$ provide complementary information that enhances the predictive performance.
    \item \textbf{Prompt Augmentation.}
    We compare the performance of \method using different in-context sampling strategies.
    As shown in (3) of Table~\ref{tab:ablation}, examples selected by our multi-modal encoder, which leverages both time series and text data, provide more meaningful demonstrations, as evidenced by its superior performance compared to random sampling and the time series-only encoder.
\end{itemize}

\noindent\textbf{Prediction Fusion.}
The final prediction of \method is obtained by aggregating the predictions from the multi-modal encoder $\mathcal{E}_\phi$ and the LLM agent $\mathcal{A}_\text{P}$. 
As shown in (4) of Table~\ref{tab:ablation}, this combined approach generally enhances overall performance, often exceeding the best performance of each model independently (i.e., select-one).
This indicates that the predictions from the two models are complementary, leveraging each other's strengths.

\subsection{Interpretability\label{sec:exp:interpretability}}
We evaluate the interpretation provided by \method (see Section~\ref{sec:method:interpretation}) through a case study, as illustrated in Figure~\ref{fig:interpretation}.

\noindent\textbf{Implicit Interpretation.}
The presence of in-context examples significantly affects interpretation and prediction.
Without in-context examples, the LLM predicts the outcome solely based on the input data, which can result in incorrect predictions. 
In contrast, with in-context examples, the LLM leverages past text-outcome relationships, leading to more informed predictions and interpretations.

\noindent\textbf{Explicit Interpretation.}
The LLM agent selects a text summary from the provided in-context examples that align with the input text. 
The selected in-context examples serve as valuable references for post-hoc interpretation of the prediction. 
Moreover, the corresponding time series can be compared with the input time series for further interpretation.

\subsection{Further Analyses\label{sec:exp:additional}}
We provide additional experimental results with \method.

\noindent\textbf{Few-Data Results.} 
We evaluate \method and two leading competitors, PatchTST and GPT4TS, on a reduced training set, specifically reducing it to 10\% of the default training ratio and even to 0\%.
As shown in Figure~\ref{fig:few}, while the competitors suffer from data scarcity, \method relatively maintains its performance with the reduced training size. 
Furthermore, it achieves high zero-shot performance, which demonstrates the effectiveness of \method in data-scarce scenarios, which is valuable for real-world applications.

\begin{table}
      \centering
      \scalebox{0.805}{
        \begin{tabular}{c@{\hspace{4.6pt}}c|c@{\hspace{4.6pt}}c@{\hspace{4.6pt}}c}
            \toprule
            Classifier & In-Context Sampler & Weather & Finance & Healthcare\\
            \midrule
            \multirow{2}{*}{KNN} & PatchTST & 0.540 & 0.325 & 0.657\\
            & MM Encoder & \textbf{0.555} & \textbf{0.338} & \textbf{0.736}\\
            \midrule
            \multirow{3}{*}{LLM ($\mathcal{A}_\text{P}$)} & None (Zero-shot) & 0.591 & 0.347 & 0.896\\
            & PatchTST & 0.598 & 0.342 & 0.948\\
            & MM Encoder & \textbf{0.615} & \textbf{0.371} & \textbf{0.954}\\
            \bottomrule
        \end{tabular}}
    \caption{\label{tab:knn} Our multi-modal (MM) encoder selects more useful in-context examples than PatchTST, as indicated by higher average F1 scores in each domain, enabling $\mathcal{A}_\text{P}$ to make more accurate event predictions. }
\end{table}

\noindent\textbf{Quality of In-Context Examples.}
We evaluate the quality of in-context examples selected by our multi-modal encoder compared to those chosen by PatchTST using a KNN classifier, which predicts the class of the input based on the majority labels of the $k$ in-context examples. 
As shown in Table~\ref{tab:knn}, the KNN with our multi-modal encoder outperforms that with PatchTST, indicating that our encoder generates embeddings that are more useful as in-context examples.
Consequently, this leads to more accurate predictions by $\mathcal{A}_\text{P}$ when used as in-context examples.

\section{Conclusion}
\label{sec:conclusions}
In this work, we introduce \method, a novel framework that leverages LLM's contextual understanding for time series event prediction.
\method employs two independent LLM agents for contextualization and prediction, supported by a trainable multi-modal encoder that mutually enhances them. 
Our experimental results on seven real-world time series datasets from various domains demonstrate the effectiveness of \method.
The datasets used in this paper are available at \url{https://github.com/geon0325/TimeCAP}. The code is available upon request.

\section*{Acknowledgements}
This work was partly supported by Institute of Information \& Communications Technology Planning \& Evaluation (IITP) grant funded by the Korea government (MSIT) (No. RS-2024-00438638, EntireDB2AI: Foundations and Software for Comprehensive Deep Representation Learning and Prediction on Entire Relational Databases, 50\%)
(No. 2019-0-00075 / RS-2019-II190075, Artificial Intelligence Graduate School Program (KAIST), 10\%).
This work was partly supported by the National Research Foundation of Korea (NRF) grant funded by the Korea government (MSIT) (No. RS-2024-00406985, 40\%).
\bibliography{ref}


\end{document}